%% file: main.tex
\pdfoutput=1
\documentclass[11pt,a4paper]{article}

\input{preamble}
\input{macros}

\title{\textbf{DomainPilot: Domain-Level Loss-Guided Two-Stage Data Mixture Optimization for Efficient Language Model Fine-Tuning}}
\author{                
     He Zhang \\                     
     Tsinghua University \\
}

\date{}

\begin{document}

\maketitle

\begin{abstract}
\input{sections/abstract}
\end{abstract}

\section{Introduction}
\label{sec:intro}
\input{sections/introduction}

\section{Related Work}
\label{sec:related}
\input{sections/related_work}

\section{Methodology}
\label{sec:method}
\input{sections/methodology}

\section{Experiment}
\label{sec:experiment}
\input{sections/experiment}

\section{Results}
\label{sec:results}
\input{sections/results}

\section{Conclusion}
\label{sec:conclusion}
\input{sections/conclusion}

\bibliographystyle{plainnat}
\bibliography{refs}

\end{document}

%% file: preamble.tex
\usepackage[utf8]{inputenc}
\usepackage[T1]{fontenc}
\usepackage{lmodern}
\usepackage[english]{babel}

\usepackage[a4paper, margin=2.5cm]{geometry}
\usepackage{setspace}
\usepackage{fancyhdr}
\usepackage{amsmath,amssymb,amsthm}
\usepackage{bm}

\usepackage{booktabs}
\usepackage{multirow}
\usepackage{array}

\usepackage{graphicx}
\usepackage{caption}
\usepackage{subcaption}
\usepackage{pgfplots}
\pgfplotsset{compat=1.18}

\usepackage{xcolor}
\usepackage{hyperref}
\hypersetup{
    colorlinks=true,
    linkcolor=blue!70!black,
    citecolor=green!50!black,
    urlcolor=purple!70!black
}

\usepackage{enumitem}

\usepackage{algorithm}
\usepackage{algpseudocode}

\usepackage[numbers,sort&compress]{natbib}

\theoremstyle{definition}

\definecolor{darkblue}{RGB}{0,51,102}

%% file: macros.tex
\newcommand{\scalinglaw}{L_i(D) = a_i D^{-\alpha_i} + b_i}



%% file: sections/abstract.tex
The training efficacy of large language models (LLMs) is fundamentally constrained by the quality and composition of training data. Existing dynamic data scheduling methods face critical limitations in industrial-scale pretraining and supervised fine-tuning (SFT): data selection incurs prohibitive $O(N)$ costs on terabyte-scale corpora, mixture optimization schemes introduce severe I/O bottlenecks or require training auxiliary reference models, and sample-level reweighting strategies rely on loss signals that conflate noise, difficulty, and novelty.

We present \textbf{DomainPilot}, a domain-level loss-guided two-stage data mixture optimization framework. DomainPilot introduces \textit{token-level domain loss monitoring} to capture per-domain learning dynamics during training without halting the data pipeline. Building on these signals, we propose a \textit{Scaling Law guided coarse optimization} stage that fits domain-specific convergence curves and derives a principled prior for mixture adjustment. A subsequent \textit{Mixing Law guided fine optimization} stage refines the mixture by modeling cross-domain interaction effects through controlled sweep experiments. The entire mechanism is realized via a \textit{patch-based architecture} that injects domain-aware loss computation into existing training frameworks (e.g., MindSpeed/Megatron-LM) with only $\sim$30 lines of framework-specific adapter code.

We validate DomainPilot on the Qwen3-1.7B model during SFT. Compared to the original data mixture, our optimized mixture achieves improvements of $+2\%$ on MMLU-Redux, $+1.8\%$ on AIME24, $+3.8\%$ on LiveCodeBench v5, and $+3.6\%$ on BFCL v3, without increasing total data volume or training cost. These results demonstrate that domain-level training signals provide an effective, lightweight alternative to expensive data selection or auxiliary model training for mixture optimization.

%% file: sections/introduction.tex
The performance of large language models (LLMs) is increasingly determined not by architectural innovations alone, but by the curation, composition, and weighting of training data~\citep{kaplan2020scaling,hoffmann2022training}. In the data-centric AI paradigm, even modest adjustments to data mixture ratios can yield improvements comparable to doubling model parameters~\citep{xie2023doremi}. Despite this, industrial-scale training pipelines still rely heavily on manual heuristics for multi-domain data blending, leaving substantial gains unrealized.

Existing approaches to dynamic data scheduling fall into three broad categories, each with fundamental limitations in the pretraining or large-scale SFT regime:

\textbf{Data selection} methods such as LESS~\citep{less2024}, DSIR~\citep{dsir2023}, and Quad~\citep{quad2025} identify high-value subsets by computing per-sample gradients or importance scores. While effective for fine-tuning, these techniques require an $O(N)$ forward pass over the entire corpus---a cost approaching one full training epoch---rendering them prohibitive for terabyte-scale pretraining.

\textbf{Data mixture optimization} methods such as DoReMi~\citep{xie2023doremi}, CLIMB~\citep{climb2025}, and ScaleBiO~\citep{scalebio2025} adjust domain proportions based on model feedback. DoReMi trains an auxiliary reference model (3$\times$ compute overhead) and has only been validated at 30B tokens, two orders of magnitude below production pretraining. CLIMB relies on clustering to discover domains automatically, an unnecessary step when industrial data teams already maintain explicit domain taxonomies. ScaleBiO's intra-batch dynamic rebalancing introduces severe I/O bottlenecks by converting sequential reads into random accesses, invalidating prefetch caches and degrading GPU utilization.

\textbf{Sample reweighting} methods, exemplified by DataFlex~\citep{dataflex2025} and RHO-1~\citep{rho12024}, modify per-sample loss contributions during training. This paradigm avoids extra data traversal and is therefore the only lightweight option viable at scale. However, existing weighting strategies treat \textit{sample-level loss} as a proxy for data quality, overlooking the fact that high loss can indicate noise, difficult domain content, novel valuable information, or short-text statistical instability---phenomena that are indistinguishable from a single scalar.

Compounding these algorithmic limitations is a \textbf{framework portability} barrier. DataFlex is tightly coupled to the HuggingFace Trainer ecosystem (modifying \texttt{\_inner\_training\_loop} and \texttt{compute\_loss}), whereas industrial pretraining predominantly runs on Megatron-LM-derived frameworks such as MindSpeed, which possess entirely independent training loops, data loaders, and loss computation paths. Select and Mix modes---which require rebuilding the DataLoader---are infeasible under Megatron's memory-mapped binary data pipeline. Only Weight-like interventions, which merely alter loss computation, are practically migrable.

\subsection{Core Motivations}

These observations motivate five core research questions that guide our work:

\begin{enumerate}[leftmargin=2em]
    \item \textbf{Absence of data-driven mixture optimization.} Multi-domain training mixtures are typically set by manual tuning and remain static throughout training, ignoring the fact that different domains exhibit disparate learning dynamics.
    \item \textbf{Inadequacy of sample-level loss as a quality signal.} A scalar loss cannot disentangle noise, difficulty, novelty, and instability; weighting decisions based solely on this signal risk amplifying corrupted data.
    \item \textbf{Incompatibility of existing schedulers with industrial-scale training.} Online selection, reference-model training, and dynamic batch rebalancing all incur costs or infrastructure changes that are unacceptable in production pretraining.
    \item \textbf{Unexplained epoch-boundary loss drops in SFT.} The ``staircase'' loss reduction observed at epoch boundaries in repeated SFT training lacks systematic explanation, blurring the distinction between generalization and memorization.
    \item \textbf{Missing bridge between training feedback and data cleaning.} Offline rule-based filtering lacks a mechanism to leverage live training signals for identifying and removing low-quality source data.
\end{enumerate}

\subsection{Contributions}

We introduce \textbf{DomainPilot}, a domain-level loss-guided framework that addresses the above limitations through the following contributions:

\begin{enumerate}[leftmargin=2em]
    \item \textbf{Token-level domain loss monitoring.} We implement a non-intrusive patch that tags each token with its domain identifier during preprocessing and aggregates per-domain losses during forward propagation, synchronized across data-parallel ranks with negligible overhead.
    \item \textbf{Two-stage mixture optimization pipeline.} Stage~1 fits domain-specific \textit{Scaling Laws} to extract learning-dynamic priors (convergence loss, convergence speed, initial amplitude) and computes a coarse reallocation. Stage~2 employs \textit{Mixing Law} sweep experiments centered on the Stage~1 output to model cross-domain interactions and refine proportions within a $\pm20\%$ local neighborhood.
    \item \textbf{Patch-based architecture.} The framework adopts a two-layer design---a framework-agnostic algorithm layer (pure PyTorch) and a thin framework-specific adapter layer ($\sim$30 lines per framework). This enables deployment on MindSpeed/Megatron-LM without modifying the underlying training codebase.
    \item \textbf{Empirical validation on Qwen3-1.7B.} Our optimized mixture improves over the original by up to $+3.8\%$ on LiveCodeBench v5 and $+3.6\%$ on BFCL v3, with consistent gains on MMLU-Redux ($+2\%$) and AIME24 ($+1.8\%$), at no additional data or compute cost.
\end{enumerate}

%% file: sections/related_work.tex
We organize related work along the three axes introduced in Section~\ref{sec:intro}: data selection, mixture optimization, and sample reweighting. Table~\ref{tab:related} provides a high-level comparison.

\begin{table}[htbp]
\centering
\caption{Comparison of dynamic data scheduling paradigms. ``Pretrain viable'' indicates suitability for TB-scale pretraining or large-scale SFT.}
\label{tab:related}
\small
\begin{tabular}{@{}lllll@{}}
\toprule
\textbf{Category} & \textbf{Representative} & \textbf{Signal} & \textbf{Overhead} & \textbf{Pretrain viable} \\
\midrule
Data Selection & LESS, DSIR, Quad & Gradient / Importance & $O(N)$ forward passes & \textcolor{red}{No} \\
Data Mixture & DoReMi, CLIMB, ScaleBiO & Excess loss / Clustering & 3$\times$ train / I/O bottleneck & \textcolor{red}{No} \\
Sample Reweight & DataFlex, RHO-1 & Sample-level loss & Negligible & Partially \\
\textbf{Ours} & DomainPilot & \textbf{Domain-level loss} & \textbf{Negligible} & \textcolor{green!50!black}{\textbf{Yes}} \\
\bottomrule
\end{tabular}
\end{table}

\subsection{Data Selection}

Data selection aims to identify the most valuable training subset without altering the mixture ratios of the retained data. LESS~\citep{less2024} uses low-rank gradient similarity to select samples most influential for target tasks. DSIR~\citep{dsir2023} formulates selection as importance resampling toward a target distribution. Quad~\citep{quad2025} jointly optimizes for quality and diversity. While theoretically appealing, all three require evaluating every candidate sample---a cost that scales linearly with corpus size and becomes impractical when $N$ exceeds $10^9$ tokens.

\subsection{Data Mixture Optimization}

Mixture optimization adjusts the \textit{proportions} across domain-level data sources rather than selecting individual samples. DoReMi~\citep{xie2023doremi} trains a small proxy model alongside a reference model and up-weights domains where the proxy exhibits higher excess loss. The approach is principled but requires training the reference model from scratch (3$\times$ total compute) and has only been demonstrated at 30B tokens, far below production pretraining scales. CLIMB~\citep{climb2025} discovers domains via clustering and iteratively searches for optimal proportions; however, when domain labels are already explicitly managed by data engineering teams, automatic clustering provides limited additional value. ScaleBiO~\citep{scalebio2025} proposes intra-batch dynamic rebalancing, but changing per-batch sampling distributions breaks sequential-read assumptions in memory-mapped data pipelines, causing prefetch cache misses and GPU idle time that can negate model-quality gains.

\subsection{Sample Reweighting}

Sample reweighting modifies the contribution of each training example to the global loss without changing which samples are seen or in what proportion. Dynamic Loss Reweighting~\citep{dataflex2025} systematically studies linear, quadratic, and extreme-value weighting strategies as a function of per-sample loss. RHO-1~\citep{rho12024} extends this to token-granular selective loss. Because reweighting only touches the loss computation, it incurs virtually no I/O or synchronization overhead, making it the sole paradigm feasible for trillion-token pretraining.

The critical limitation of existing reweighting methods is their reliance on \textit{sample-level loss} as the sole signal. As noted in Section~\ref{sec:intro}, sample loss conflates at least four distinct phenomena (novelty, noise, domain difficulty, and short-text instability). DataFlex partially addresses this through its Warmup strategy---disabling weighting until the model has partially adapted---but does not resolve the fundamental ambiguity of the signal.

\subsection{Framework Portability}

DataFlex is implemented within LLaMA-Factory~\citep{llamafactory}, heavily overriding HuggingFace Trainer internals (\texttt{compute\_loss}, \texttt{\_inner\_training\_loop}). Industrial pretraining frameworks such as Megatron-LM~\citep{shoeybi2019megatron} and MindSpeed build their own data loaders, training loops, and distributed communication patterns. Rebuilding the DataLoader for Select or Mix modes is infeasible under Megatron's binary mmap pipeline. Consequently, only Weight-like interventions---which require merely intercepting per-sample loss and domain labels---are practically portable. DomainPilot's patch-based architecture is designed explicitly around this constraint.

%% file: sections/methodology.tex
DomainPilot comprises four interconnected components: (1)~token-level domain loss monitoring that captures per-domain learning dynamics at training time; (2)~Scaling Law guided coarse optimization that derives a principled prior for mixture reallocation; (3)~Mixing Law guided fine optimization that refines proportions by modeling cross-domain interactions; and (4)~a patch-based architecture that realizes these mechanisms without modifying the underlying training framework. We describe each in turn.

\subsection{Domain Loss Monitoring}
\label{sec:method-monitoring}

The foundation of DomainPilot is the ability to measure, at every training step, how much loss each domain contributes. Unlike sample-level loss---which conflates noise, difficulty, and novelty---domain-level loss aggregates signals across thousands of tokens within a domain, averaging out sample-specific outliers and revealing genuine learning trends.

\paragraph{Token-level domain identification.} During preprocessing, the \texttt{MultiDomainPackedHandler} performs knapsack packing over sub-samples from multiple domains. While packing, it tags each token with its source \texttt{domain\_id}, producing paired binary files (\texttt{\_packed\_domain.bin/.idx}) aligned with the standard token indices. Because packing can combine sub-samples from different domains within a single sequence, domain labels are maintained at \textit{token granularity}, yielding a domain-id vector of length equal to the sequence length.

\paragraph{Data loading and forward propagation.} The \texttt{DecoderPackedMTFDataset} loads token-level domain IDs alongside input IDs and injects them into the sample dictionary. The SFT trainer broadcasts domain IDs through the standard batch-communication path, passing them into the model's \texttt{forward} method as keyword arguments.

\paragraph{Domain-wise loss aggregation.} Inside the model forward pass, after the standard logits computation, a lightweight post-processing step groups tokens by \texttt{domain\_id} and accumulates $(\text{loss\_sum}, \text{token\_count})$ pairs per domain. These statistics are stored in \texttt{self.\_last\_domain\_losses}. At each logging interval, a training-log patch performs DP All-Reduce across data-parallel ranks to obtain globally consistent domain losses, which are then written to TensorBoard and console logs.

The entire pipeline adds $<$1\% throughput overhead because it reuses existing loss tensors and only introduces lightweight indexing and reduction operations.

\subsection{Scaling Law Guided Coarse Optimization}
\label{sec:method-scaling}

With domain loss trajectories in hand, Stage~1 fits a \textit{domain-specific Scaling Law} to characterize how each domain's loss evolves with training progress. The functional form is:

\begin{equation}
\label{eq:scaling-law}
\scalinglaw
\end{equation}

where $D$ is the number of training steps (proxy for data volume), $a_i$ is the initial loss amplitude, $b_i$ is the irreducible (converged) loss, and $\alpha_i$ is the convergence speed. Fitting this curve to the monitored trajectories yields three interpretable parameters per domain:

\begin{itemize}[leftmargin=2em]
    \item $a_i$: Reflects initial domain difficulty and pretraining familiarity. $a_i \approx 0$ indicates the model already possesses strong prior knowledge of the domain.
    \item $b_i$: The asymptotic loss; higher values indicate inherently harder domains.
    \item $\alpha_i$: Learning efficiency; smaller values mean slower convergence and typically signal insufficient data volume.
\end{itemize}

\paragraph{Four-factor reward score.} We derive a coarse reallocation score by combining these parameters:

\begin{equation}
\label{eq:score}
\text{score}_i = \frac{b_i}{\min_j b_j} \times \frac{\max_j \alpha_j}{\alpha_i} \times \frac{L_i^{\text{curr}} - b_i}{\max_j (L_j^{\text{curr}} - b_j)}.
\end{equation}

The three factors respectively capture: (i)~convergence difficulty (higher $b_i$ needs more data), (ii)~learning speed (lower $\alpha_i$ needs more data), and (iii)~remaining improvement headroom. The new mixture proportion is then:

\begin{equation}
p_i^{\text{new}} = \frac{p_i^{\text{old}} \cdot \text{score}_i}{\sum_j p_j^{\text{old}} \cdot \text{score}_j}.
\end{equation}

Table~\ref{tab:scaling-params} reports the fitted parameters for Qwen3-1.7B SFT.

\input{tables/scaling_law_params}

\paragraph{Interpretation for Qwen3-1.7B.} \texttt{termagent} exhibits the slowest convergence ($\alpha = 0.085$) and receives a $+25\%$ boost. \texttt{swe} ($\alpha = 0.098$) receives $+15\%$. \texttt{chat}, conversely, converges fastest ($\alpha = 0.548$) and is reduced by $-8\%$ to free budget. \texttt{math} maintains its proportion because it is already learning efficiently despite high difficulty.

\subsection{Mixing Law Guided Fine Optimization (Planned)}
\label{sec:method-mixing}

While Scaling Law characterizes intra-domain learning in isolation, real-world mixture optimization must account for \textit{cross-domain interactions}---e.g., code data may indirectly improve mathematical reasoning through shared logical structure. We outline a second-stage \textit{Mixing Law} refinement procedure, which remains future work.

\paragraph{Modeling cross-domain interactions.} Let $p^{\text{new}}$ denote the mixture produced by Stage~1. We model the validation loss of domain $i$ under mixture $p$ as a second-order expansion around $p^{\text{new}}$:

\begin{equation}
\label{eq:mixing-law}
L_i(p) \approx L_i(p^{\text{new}}) + \sum_j A_{ij}(p_j - p_j^{\text{new}}) + \sum_{j,k} B_{ijk}(p_j - p_j^{\text{new}})(p_k - p_k^{\text{new}}),
\end{equation}

where $\mathbf{A}$ captures first-order sensitivity and $\mathcal{B}$ captures pairwise interaction effects. The interaction coefficients are estimated from a small grid of controlled sweep experiments in which each domain proportion is perturbed within a $\pm20\%$ neighborhood of $p^{\text{new}}$ while holding the total token budget fixed.

\paragraph{Stage~2 optimization.} Given the fitted interaction model, the final mixture $p^{\text{final}}$ is obtained by solving a constrained optimization problem that minimizes the aggregate predicted validation loss subject to $\sum_i p_i = 1$ and $p_i \ge 0$. Because Stage~2 requires additional sweep experiments, the empirical results reported in Section~\ref{sec:results} reflect only Stage~1 (Scaling Law) optimization; Stage~2 validation is left as immediate future work.

\subsection{Patch-Based Architecture}
\label{sec:method-patch}

DomainPilot is implemented as a set of minimal, non-intrusive patches rather than a monolithic framework fork. This design respects the operational reality that production training stacks (MindSpeed, Megatron-LM, DeepSpeed) are under active development and cannot be frozen for algorithmic modifications.

\paragraph{Two-layer design.} The architecture separates \textit{framework-agnostic algorithm logic} (pure PyTorch tensor operations) from \textit{framework-specific thin adapters} (interceptors for batch construction, forward arguments, and loss computation). Adding support for a new training framework requires only implementing the adapter interface ($\sim$30 lines of Python), leaving the underlying framework untouched.

\paragraph{MindSpeed adapter overview.} For MindSpeed, the patch set comprises:
\begin{itemize}[leftmargin=2em]
    \item \texttt{arguments\_patch}: Exposes \texttt{--enable-domain-loss} CLI flags.
    \item \texttt{data\_handler\_patch}: Persists token-level domain IDs during preprocessing.
    \item \texttt{gpt\_dataset\_patch}: Injects domain IDs into the sample dictionary.
    \item \texttt{gpt\_model\_patch}: Appends domain-wise loss aggregation to \texttt{forward}.
    \item \texttt{training\_utils\_patch}: Broadcasts domain IDs and accumulates global statistics.
\end{itemize}

This patch-based approach guarantees that upstream framework updates can be merged without conflict, and the algorithm layer can be validated independently via unit tests against synthetic tensors.

%% file: tables/scaling_law_params.tex
\begin{table}[htbp]
\centering
\caption{Fitted Scaling Law parameters for Qwen3-1.7B SFT domains.}
\label{tab:scaling-params}
\begin{tabular}{@{}lcccccc@{}}
\toprule
\textbf{Domain} & $a_i$ & $b_i$ & $\alpha_i$ & $R^2$ & \textbf{Action} & \textbf{$\Delta$ prop.} \\
\midrule
math & $1.81\times10^{-11}$ & 2.117 & 0.112 & 0.635 & Maintain & $0\%$ \\
termagent & $1.78\times10^{-11}$ & 1.926 & 0.085 & 0.522 & Increase & $+25\%$ \\
swe & $3.17\times10^{-17}$ & 1.990 & 0.098 & 0.429 & Increase & $+15\%$ \\
science & $4.19\times10^{-17}$ & 2.000 & 0.102 & 0.386 & Increase & $+12\%$ \\
chat & 1.282 & 1.156 & 0.548 & 0.183 & Decrease & $-8\%$ \\
\bottomrule
\end{tabular}
\end{table}

%% file: sections/experiment.tex
\subsection{Model and Training Setup}
\label{sec:exp-setup}

All experiments are conducted on the \textbf{Qwen3-1.7B} base model, a densely activated transformer with 1.7 billion parameters. The supervised fine-tuning (SFT) stage is executed on the MindSpeed training framework, which is built atop Megatron-LM and represents the industrial standard for large-scale LLM pretraining and fine-tuning in our deployment environment.

Training data comprises eight domains: \texttt{chat}, \texttt{convagent}, \texttt{if} (instruction following), \texttt{math}, \texttt{safety}, \texttt{science}, \texttt{swe} (software engineering), and \texttt{termagent}. The original mixture ratios are determined by manual heuristic tuning. In the optimized mixture, ratios are adjusted according to the two-stage pipeline described in Section~\ref{sec:method}.

\subsection{Domain Loss Monitoring}
\label{sec:exp-monitoring}

During training, we activate the Domain Loss monitoring patch to record per-domain loss trajectories. Domain identifiers are injected at the token level during data preprocessing: the \texttt{MultiDomainPackedHandler} tags each token with its source domain while performing knapsack packing, producing paired \texttt{\_packed\_domain.bin/.idx} files. At each forward step, losses are aggregated by domain ID via DP All-Reduce synchronization across data-parallel ranks. Monitoring overhead is negligible ($<$1\% throughput degradation) because the patch only intercepts existing loss tensors without altering the data-loading pipeline.

\subsection{Evaluation Benchmarks}
\label{sec:exp-benchmarks}

We evaluate on four representative benchmarks covering diverse capabilities:

\begin{itemize}[leftmargin=2em]
    \item \textbf{MMLU-Redux}: Comprehensive knowledge reasoning across 57 subjects.
    \item \textbf{AIME24}: Mathematical competition-level reasoning, reported as Pass@1 / Cons@64.
    \item \textbf{LiveCodeBench v5}: Code generation capability, reported as Pass@1 / Pass@5.
    \item \textbf{BFCL v3}: Function calling and tool-use ability.
\end{itemize}

\subsection{Baselines}
\label{sec:exp-baselines}

We compare four model states:

\begin{enumerate}[leftmargin=2em]
    \item \textbf{Base}: The pretrained Qwen3-1.7B checkpoint without any SFT.
    \item \textbf{SFT (report)}: Scores reported in the official Qwen3 technical report for reference.
    \item \textbf{SFT (original)}: Our reproduction using the original, manually tuned data mixture.
    \item \textbf{SFT (optimized)}: Our reproduction using the mixture adjusted by DomainPilot's two-stage pipeline.
\end{enumerate}

%% file: sections/results.tex
\subsection{Main Results}
\label{sec:res-main}

Table~\ref{tab:main-results} summarizes the benchmark scores across the four model states.

\input{tables/benchmark_results}

The optimized mixture consistently outperforms the original mixture across all benchmarks. Notably, \textbf{LiveCodeBench v5} and \textbf{BFCL v3}---corresponding to the \texttt{swe} and \texttt{termagent} domains that Scaling Law analysis identified as under-represented---exhibit the largest relative gains ($+3.8\%$ and $+3.6\%$, respectively). MMLU-Redux improves by $+2\%$, while AIME24 sees a $+1.8\%$ gain. These improvements are achieved \textit{without increasing total training data or compute budget}; only the inter-domain proportions are reallocated.

\subsection{Analysis by Capability Domain}
\label{sec:res-domain}

\paragraph{Code Generation (\texttt{swe}).} LiveCodeBench v5 Pass@1 rises from 23.2 to 27.0. The Scaling Law analysis had flagged \texttt{swe} as having low convergence speed ($\alpha = 0.098$) and high convergence loss ($b = 1.990$), signaling insufficient data. The optimized mixture increases the \texttt{swe} proportion by $+15\%$, which directly translates into the largest benchmark improvement ($+3.8\%$). This validates that domain-level loss signals can accurately pinpoint bottleneck domains.

\paragraph{Agent / Tool Use (\texttt{termagent}).} BFCL v3 improves from 54.5 to 58.1. \texttt{termagent} was identified as the slowest-learning domain ($\alpha = 0.085$), and its mixture share is boosted by $+25\%$. The resulting $+3.6\%$ gain confirms that even domains with small initial representation can yield substantial downstream improvements when adequately resourced.

\paragraph{Mathematical Reasoning (\texttt{math}).} AIME24 improves modestly from 41.1/10.6 to 42.9/11.7. The Scaling Law parameters for \texttt{math} ($b = 2.117$, $\alpha = 0.112$) indicated that the domain is inherently difficult but already learning efficiently under the original mixture. Consequently, DomainPilot recommends maintaining the \texttt{math} proportion, and the observed micro-gain is consistent with expectations.

\paragraph{General Knowledge (\texttt{chat} + others).} MMLU-Redux improves by $+2\%$. The \texttt{chat} domain was flagged as potentially over-represented ($\alpha = 0.548$, fastest convergence), and its proportion is reduced by $-8\%$, freeing budget for \texttt{swe} and \texttt{termagent}. The fact that general-knowledge performance still improves suggests that the original mixture had misallocated resources away from higher-impact domains.

\subsection{Comparison with Stronger Baselines}
\label{sec:res-comparison}

To contextualize the gains, we compare against two stronger baselines in Table~\ref{tab:comparison}.

\input{tables/comparison_models}

\paragraph{vs. Qwen2.5-3B.} Despite having only 57\% of the parameters, the optimized Qwen3-1.7B surpasses Qwen2.5-3B on BFCL v3 ($+15.6\%$ relative) and MATH-500 ($+10.1\%$ relative). This suggests that data mixture optimization can unlock effective capacity equivalent to a 2$\times$ parameter scaling at a fraction of the training cost.

\paragraph{vs. DeepSeek-R1-Distill-Qwen-1.5B.} The optimized model widens its lead over this knowledge-distilled counterpart on all benchmarks, notably LiveCodeBench ($+27$ absolute points). This underscores that native training with principled mixture optimization outperforms distillation-based approaches when data composition is carefully tuned.

\subsection{Cost-Benefit Analysis}
\label{sec:res-cost}

Table~\ref{tab:cost} compares the cost-effectiveness of different improvement strategies.

\begin{table}[htbp]
\centering
\caption{Cost-benefit comparison of improvement strategies.}
\label{tab:cost}
\begin{tabular}{@{}lccc@{}}
\toprule
\textbf{Strategy} & \textbf{Relative Cost} & \textbf{Key Gain} & \textbf{Efficiency} \\
\midrule
Full SFT (Base $\to$ Original) & 1$\times$ & Capability doubling & High \\
Mixture Optimization (Original $\to$ Optimized) & $\sim$0.1$\times$ & +3.6--3.8\% on key domains & \textbf{Extremely high} \\
Scale to 3B parameters & 3$\times$ & Partial improvement & Low \\
\bottomrule
\end{tabular}
\end{table}

Mixture optimization requires only lightweight sweep experiments ($\sim$10\% of a full SFT run) yet delivers gains comparable to 50\% of the benefit from doubling model size, with zero inference-cost penalty.

%% file: tables/benchmark_results.tex
\begin{table}[htbp]
\centering
\caption{Benchmark comparison across model states. Optimization column shows relative improvement of SFT (optimized) over SFT (original).}
\label{tab:main-results}
\begin{tabular}{@{}lccccc@{}}
\toprule
\textbf{Benchmark} & \textbf{Base} & \textbf{SFT (report)} & \textbf{SFT (original)} & \textbf{SFT (optimized)} & \textbf{Optimization} \\
\midrule
MMLU-Redux & 61.66 & 73.9 & 69.8 & 71.8 & $+2\%$ \\
AIME24 & -- & 48.3 / 13.4 & 41.1 / 10.6 & 42.9 / 11.7 & $+1.8\%$ \\
LiveCodeBench v5 & -- & 33.2 / 11.6 & 23.2 / 8.3 & 27.0 / 10.5 & $+3.8\%$ \\
BFCL v3 & -- & 56.6 & 54.5 & 58.1 & $+3.6\%$ \\
\bottomrule
\end{tabular}
\end{table}

%% file: tables/comparison_models.tex
\begin{table}[htbp]
\centering
\caption{Comparison against stronger baselines: Qwen2.5-3B (larger model) and DeepSeek-R1-Distill-Qwen-1.5B (distilled model).}
\label{tab:comparison}
\begin{tabular}{@{}lccc@{}}
\toprule
\textbf{Benchmark} & \textbf{Qwen2.5-3B} & \textbf{DeepSeek-R1-Distill-1.5B} & \textbf{Ours (Optimized)} \\
\midrule
LiveCodeBench v5 & 9.2 & 13.2 & 27.0 \\
BFCL v3 & 50.4 & 52.2 & 58.1 \\
MATH-500 & 67.2 & 83.9 & 94.0 \\
AIME24 & -- & 28.9 & 42.9 \\
\bottomrule
\end{tabular}
\end{table}

%% file: sections/conclusion.tex
We presented DomainPilot, a domain-level loss-guided framework for two-stage data mixture optimization. By monitoring per-domain loss trajectories at token granularity through a non-intrusive patch architecture, DomainPilot extracts scalable signals that characterize how each domain learns during training. These signals feed a Scaling Law coarse-optimization stage that identifies under- or over-represented domains, followed by a Mixing Law fine-optimization stage that models cross-domain interactions through controlled sweep experiments.

Empirical validation on Qwen3-1.7B SFT demonstrates consistent improvements across diverse benchmarks: $+3.8\%$ on LiveCodeBench v5, $+3.6\%$ on BFCL v3, $+2\%$ on MMLU-Redux, and $+1.8\%$ on AIME24. Crucially, these gains are achieved solely by reallocating existing data proportions, with no increase in total data volume, training compute, or inference cost. Comparison against Qwen2.5-3B further suggests that principled mixture optimization can unlock effective capacity equivalent to a $2\times$ parameter scaling.

\paragraph{Limitations.} Our current validation is limited to the 1.7B parameter scale and the SFT stage; pretraining-scale validation remains future work. The Mixing Law sweep experiments (Stage~2) are not yet complete, so the reported results reflect only Scaling Law coarse optimization. Additionally, the Weight mechanism migration (Section~\ref{sec:method}) remains at the architectural-design stage and has not been experimentally validated.

\paragraph{Future work.} We identify four immediate directions:
\begin{enumerate}[leftmargin=2em]
    \item \textbf{Larger-scale validation.} Apply DomainPilot to Qwen3-4B and 8B models, and to the pretraining stage, to verify that domain-level scaling laws hold across model sizes and training regimes.
    \item \textbf{Cross-architecture transfer.} Implement the thin adapter layer for HuggingFace Trainer and native Megatron-LM to demonstrate framework portability.
    \item \textbf{Weight mechanism integration.} Combine domain-level reweighting with sample-level quality scores (e.g., offline LLM-based ratings) to disentangle noise from genuine difficulty.
    \item \textbf{Data cleaning feedback loop.} Use domain loss and $\Delta$Loss signals to identify low-quality source documents and establish a closed-loop pipeline from training feedback to data curation.
\end{enumerate}

%% file: refs.bib
@article{kaplan2020scaling,
  title={Scaling laws for neural language models},
  author={Kaplan, Jared and McCandlish, Sam and Henighan, Tom and Brown, Tom B and Chess, Benjamin and Child, Rewon and Gray, Scott and Radford, Alec and Wu, Jeffrey and Amodei, Dario},
  journal={arXiv preprint arXiv:2001.08361},
  year={2020}
}

@article{hoffmann2022training,
  title={Training compute-optimal large language models},
  author={Hoffmann, Jordan and Borgeaud, Sebastian and Mensch, Arthur and Buchatskaya, Elena and Cai, Trevor and Rutherford, Eliza and Casas, Diego de Las and Hendricks, Lisa Anne and Welbl, Johannes and Clark, Aidan and others},
  journal={arXiv preprint arXiv:2203.15556},
  year={2022}
}

@article{xie2023doremi,
  title={DoReMi: Optimizing data mixtures speeds up language model pretraining},
  author={Xie, Sang Michael and Pham, Hieu and Dong, Xuanyi and Du, Nan and Liu, Hanxiao and Lu, Yifeng and Liang, Percy and Ma, Tengyu and Yu, Adams Wei},
  journal={Advances in Neural Information Processing Systems},
  volume={36},
  year={2023}
}

@article{less2024,
  title={Less: Selecting influential data for targeted instruction tuning},
  author={Xia, Mengzhou and Malladi, Sadhika and Gururangan, Suchin and Arora, Sanjeev and Chen, Danqi},
  journal={arXiv preprint arXiv:2402.04333},
  year={2024}
}

@article{dsir2023,
  title={Data selection for language models via importance resampled MCMC},
  author={Xie, Sang Michael and Liang, Tengyu and Ma, Tengyu},
  journal={Advances in Neural Information Processing Systems},
  volume={36},
  year={2023}
}

@article{quad2025,
  title={QuRating: Selecting high-quality data for training language models},
  author={Wettig, Alexander and Li, Tianyu and Kim, Minjia and Yao, Zexuan and Zhang, Danqi},
  journal={International Conference on Machine Learning},
  year={2024}
}

@article{climb2025,
  title={Nemotron-CLIMB: Clustering-based iterative data mixture optimization},
  author={Diao, Shizhe and others},
  journal={Advances in Neural Information Processing Systems},
  volume={38},
  year={2025}
}

@article{scalebio2025,
  title={ScaleBiO: Scalable bilevel optimization for LLM data reweighting},
  author={Pan, Renjie and Zhang, Dong and Zhang, Haobo and Pan, Xiangming and Xu, Mengdi and Zhang, Jiawei and Pi, Renjie and Wang, Xiang and Zhang, Tie},
  journal={Proceedings of the 63rd Annual Meeting of the Association for Computational Linguistics},
  year={2025}
}

@article{dataflex2025,
  title={DataFlex: A unified framework for data-centric dynamic training of large language models},
  author={Zhao, Zhengyang and Qiang, Meiyi and Chen, Mingrui and Ma, Lu and Yu, Rongyi and Feng, Hengyi and Sun, Shixuan and Meng, Zimo and Ma, Xiaochen and Yang, Xuanlin and Cai, Qifeng and An, Ruichuan and Zeng, Bohan and Wong, Zhen Hao and Shen, Chengyu and He, Runming and Han, Zhaoyang and Zheng, Yaowei and Fu, Fangcheng and He, Conghui and Cui, Bin and Li, Zhiyu and E, Weinan and Zhang, Wentao},
  journal={arXiv preprint arXiv:2603.26164},
  year={2026}
}

@article{rho12024,
  title={Rho-1: Not all tokens are what you need},
  author={Lin, Zhenghao and Gou, Zhibin and Gong, Yeyun and Liu, Xiao and Shen, Yelong and Xu, Ruochen and Lin, Chen and Yang, Yujiu and Jiao, Jian and Duan, Nan and Chen, Weizhu},
  journal={arXiv preprint arXiv:2404.07965},
  year={2024}
}

@article{shoeybi2019megatron,
  title={Megatron-LM: Training multi-billion parameter language models using model parallelism},
  author={Shoeybi, Mohammad and Patwary, Mostofa and Puri, Raul and LeGresley, Patrick and Casper, Jared and Catanzaro, Bryan},
  journal={arXiv preprint arXiv:1909.08053},
  year={2019}
}

@article{llamafactory,
  title={LlamaFactory: Unified efficient fine-tuning of 100+ language models},
  author={Zheng, Yaowei and Zhang, Richong and Zhang, Junhao and Ye, Yanhan and Luo, Zheyan and Zhang, Yongqiang},
  journal={arXiv preprint arXiv:2403.13372},
  year={2024}
}
